\documentclass[10pt,twocolumn,letterpaper]{article}

\PassOptionsToPackage{dvipsnames,table}{xcolor}
\usepackage[pagenumbers]{cvpr} 

\definecolor{cvprblue}{rgb}{0.21,0.49,0.74}
\usepackage[pagebackref,breaklinks,colorlinks,allcolors=cvprblue]{hyperref}
\usepackage{multirow}
\def\paperID{3324} 
\def\confName{CVPR}
\def\confYear{2026}

\title{Combating Visual Neglect and Semantic Drift in Large Multimodal Models for Enhanced Cross-Modal Retrieval}

\author{Guosheng Zhang$^*$, Linkai Liu$^*$, Keyao Wang, Haixiao Yue, Zhiwen Tan, Xiao Tan\\
Baidu Inc.\\
{\tt\small \{zhangguosheng, liulinkai, wangkeyao, yuehaixiao, tanzhiwen, tanxiao01\}@baidu.com}
}

\begin{document}

\twocolumn[{%
\renewcommand\twocolumn[1][]{#1}%
\maketitle

\begin{center}
    \centering
    \includegraphics[width=0.98\linewidth]{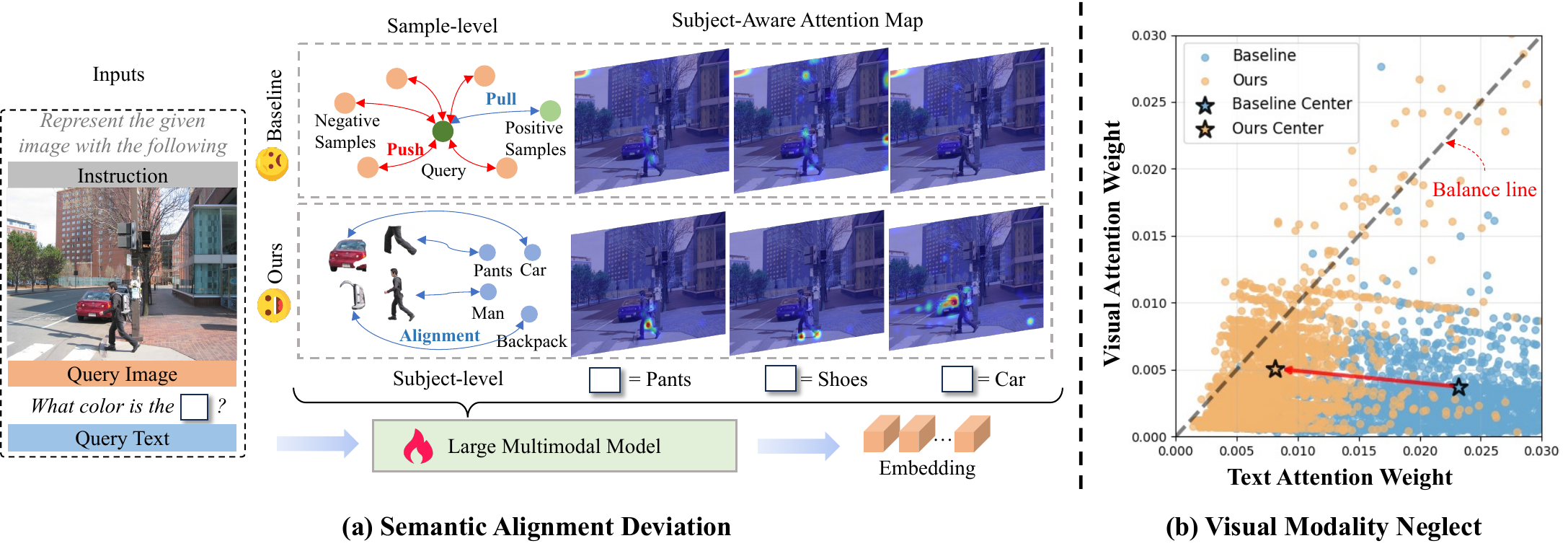}
    \vspace{-7pt}
    \captionof{figure}{The figure demonstrates two key limitations in multimodal embedding models. In (a), we observe significant semantic alignment deviation in the baseline model, which fails to accurately localize text-referred subjects. In contrast, our SSA-ME method successfully focuses on the semantically relevant visual subjects. The scatter plot in (b) reveals the visual modality neglect problem, where baseline models cluster (blue points) indicating excessive textual attention at the expense of visual information. Our method achieves a more balanced attention distribution between visual and textual modalities, as evidenced by the centered cluster distribution.}
    \label{fig:main}
\end{center}
}]
\maketitle
{\renewcommand{\thefootnote}{}\footnotetext{$^*$Equal contribution.}}
{\renewcommand{\thefootnote}{}\footnotetext{Work done during Linkai Liu's internship at Baidu Inc.}}

\begin{abstract}
Despite significant progress in Unified Multimodal Retrieval (UMR) powered by Large Multimodal Models (LMMs), existing embedding methods primarily focus on sample-level objectives via contrastive learning while overlooking the the crucial subject-level semantics. This limitation hinders the model’s ability to group semantically coherent subjects in complex multimodal queries, manifesting as \textbf{semantic alignment deviation}—where models fail to accurately localize salient text-referred regions in visual content. Moreover, without explicit guidance to model salient visual subjects, LMMs tend to over-rely on textual cues, resulting in \textbf{visual modality neglect} and suboptimal utilization of visual knowledge. To this end, we propose \textbf{S}alient \textbf{S}ubject-\textbf{A}ware \textbf{M}ultimodal \textbf{E}mbedding (SSA-ME), a novel framework designed to enhance fine-grained representation learning through saliency-aware modeling. SSA-ME leverages LMMs and visual experts to identify and emphasize salient visual concepts in image-text pairs, and introduces a saliency-guided objective to better align cross-modal attention with semantically meaningful regions. Additionally, a feature regeneration module recalibrates visual features based on the derived saliency maps, ensuring a balanced and semantically coherent integration across modalities.
Extensive experiments show that our method achieves state-of-the-art performance on the MMEB benchmark, demonstrateting that incorporating subject-level modeling substantially improves multimodal retrieval. Comprehensive qualitative analyses further illustrate the interpretability and effectiveness of our approach.
\end{abstract}
    
\section{Introduction}
\label{sec:intro}
Universal Multimodal Retrieval (UMR)~\cite{zhang2024gme, jiang2024vlm2vec, liu2025lamra, jiang2024e5, hou2023improving, robinson2020contrastive, lan2025llave, awasthi2022more} has emerged as a critical paradigm for cross-modal information systems, enabling unified representation learning across diverse modalities through Large Multimodal Models (LMMs). While existing methods have made significant progress by leveraging contrastive learning at the sample level, their predominant focus on instance discrimination overlooks the essential need for subject-level semantic understanding.This fundamental limitation becomes particularly apparent when handling complex multimodal queries that require fine-grained correspondence between visual subjects and textual references.

As illustrated in Figure~\ref{fig:main} (a), we observe significant \textbf{semantic alignment deviation} in existing methods. When processing queries involving different textual subjects, baseline models fail to establish precise visual-textual correspondences, often misallocating attention to semantically irrelevant regions rather than focusing on the actual subjects mentioned in the text. Moreover, the lack of explicit mechanisms to guide models toward salient visual subjects leads to a systematic over-reliance on textual cues—a phenomenon we term \textbf{visual modality neglect}~\cite{li2025dyfo,fazli2025mitigating}. As shown in Figure~\ref{fig:main} (b), baseline embeddings cluster predominantly in text-biased regions, indicating that multimodal representations are constructed largely from textual information while visual content remains underutilized. This imbalance fundamentally limits the model's capacity to leverage complementary visual information effectively.
These interconnected limitations—semantic misalignment due to insufficient subject-level modeling and modality imbalance arising from the lack of explicit visual subject guidance—motivate our investigation into a more fine-grained approach that explicitly incorporates subject-aware reasoning into multimodal representation learning.

In this paper, we propose Salient Subject-Aware Multimodal Embedding (SSA-ME), a novel framework that enhances multimodal representation learning at the subject level through saliency-aware mechanisms. First, we introduce a \textbf{Saliency-Guided Attention Alignment (SGA)} mechanism that leverages the powerful multimodal understanding capability of LMMs combined with visual expert models' object perception to precisely localize salient subjects in image-text pairs. A saliency-guided loss function is designed to distill the saliency detection capability into our model through latent attention alignment, thereby significantly improving the semantic correlation between visual and textual modalities. Furthermore, we develop a \textbf{Saliency-Driven Feature Regeneration (SDR)} mechanism that selectively recycles visual features based on the model's predicted saliency maps and fuses them with the global multimodal embedding. This approach fully exploits salient core features in the visual modality, enhancing the model's attention weight on visual inputs. As demonstrated in Fig.~\ref{fig:main} (a), our method accurately localizes text-mentioned salient subjects in images. Moreover, Fig.~\ref{fig:main} (b) shows that the attention distribution shifts toward the central balance line, indicating improved equilibrium between visual and textual modalities.

In summary, our contributions are threefold: 
\begin{enumerate}[label=(\arabic*)]
    \item We introduce a subject-level perspective for analyzing multimodal embedding learning, through which we identify and systematically examine two critical yet understudied limitations: semantic alignment deviation and visual modality neglect.
    \item We propose two simple yet effective components: SGA and SDR. These components enable the model to better focus on salient visual regions and mitigate visual modality neglect.
    \item Our approach achieves state-of-the-art performance on the MMEB benchmark. Experimental results confirm that explicitly modeling salient subjects significantly improves retrieval performance, demonstrating the practical value of our subject-level perspective.
\end{enumerate}

\begin{figure*}[ht!]
  \centering
   \includegraphics[width=0.99\linewidth]{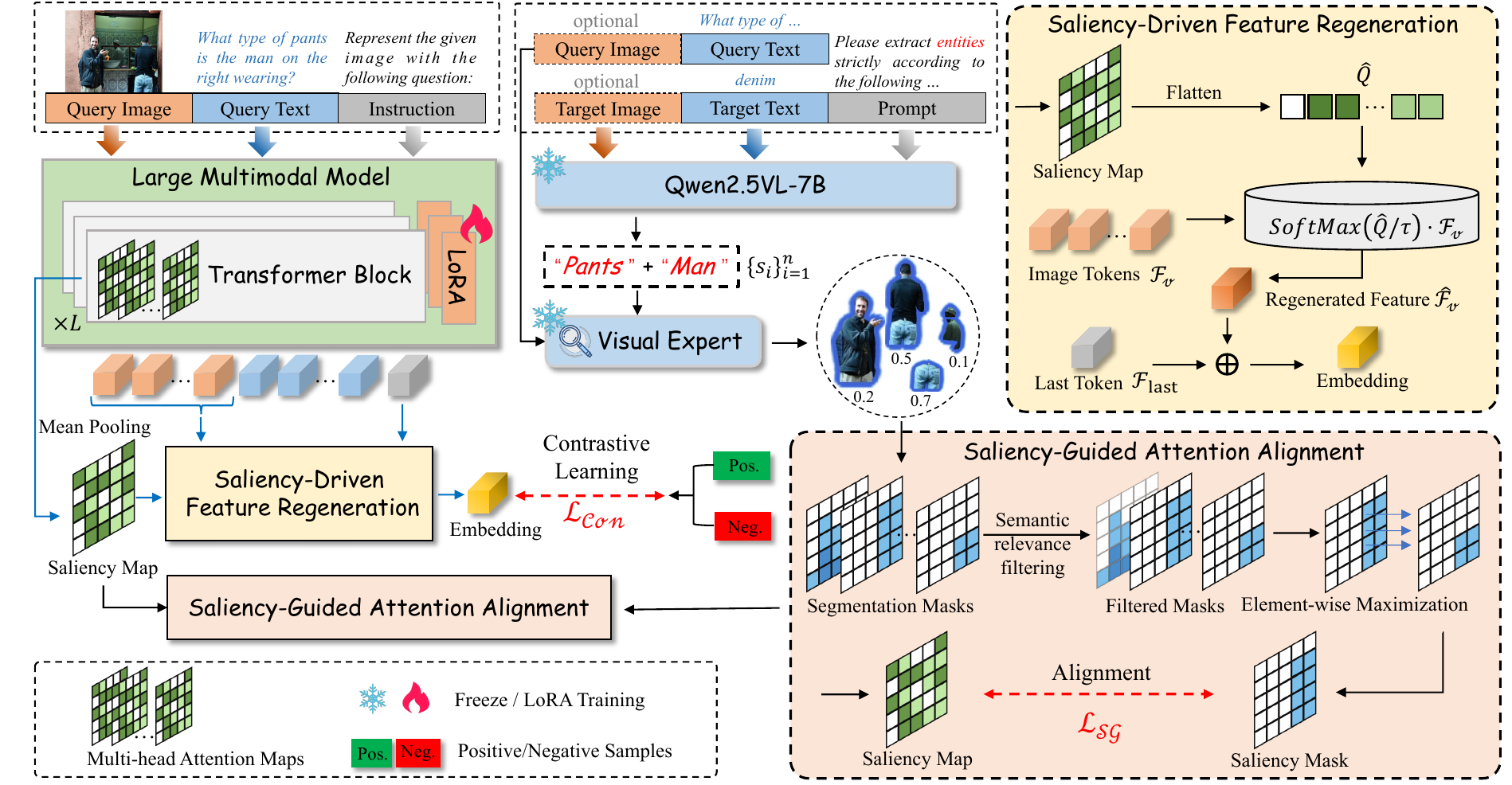}
   \caption{Overview of Salient Subject-Aware Multimodal Embedding (SSA-ME) framework. Black and blue arrows represent the training flow, while only blue arrows indicate the inference flow. To ensure the extracted salient subjects are semantically representative, we feed the query pair along with its corresponding positive sample into Qwen2.5-VL-7B~\cite{bai2025qwen2}, leveraging a carefully designed prompt to guarantee that the identified salient subjects align meaningfully with the content of the positive sample.}
   \label{fig:method}
\end{figure*}

\section{Related Works}
\textbf{Universal Multimodal Retrieval.}
Early approaches to multimodal retrieval primarily focused on unimodal settings, whereas contemporary multimedia applications increasingly demand systems capable of processing hybrid image-text inputs and generating corresponding multimodal responses. A common pipeline involves leveraging pre-trained vision-language models such as CLIP~\cite{radford2021learning} and BLIP~\cite{li2022blip} for feature extraction, combined with fusion modules as seen in UniVL-DR~\cite{liu2022universal} and UniIR~\cite{wei2024uniir}. Despite their effectiveness, these methods struggle in complex retrieval scenarios such as composed image retrieval~\cite{liu2021image,wu2021fashion,baldrati2023zero,vaze2023genecis}, long-text-to-image retrieval~\cite{zhang2024long}, and image-question-to-multimodal document retrieval~\cite{chen2023can,hu2023open}.
The field has since shifted toward leveraging Large Multimodal Models (LMMs) for their superior cross-modal understanding. Methods such as E5-V~\cite{jiang2024e5} employ carefully designed prompts to project images and text into a unified latent space, achieving strong zero-shot retrieval via text-pair fine-tuning~\cite{gao2021simcse}. GME~\cite{zhang2024gme} pioneered unified retrieval for visual documents and introduced a data synthesis pipeline for generating large-scale multimodal training data. More recently, VLM2Vec~\cite{jiang2024vlm2vec} advanced the state-of-the-art on the MMEB benchmark by fine-tuning Qwen2-VL~\cite{wang2024qwen2} with Explicit One-token Limitation (EOL), while LLaVE~\cite{lan2025llave} enhanced embedding discrimination for hard negative samples. Nevertheless, these approaches remain constrained by sample-level optimization through positive-negative contrastive learning, leaving the potential of subject-level multimodal embedding largely unexplored.
\\\textbf{Fine-Grained Visual Understanding in LMM}
Enabling LMMs to achieve fine-grained visual understanding has emerged as a critical research direction. Recent studies highlight persistent limitations in their ability to establish detailed visual-textual correspondences~\cite{kang2025see, jung2025visual,gao2025interleaved,li2025dyfo,fazli2025mitigating}. Kang et al.~\cite{kang2025see} identified the “visual attention sink” phenomenon, wherein models allocate excessive attention to irrelevant visual regions, undermining fine-grained semantic alignment. Subsequent methods, such as ICoT~\cite{gao2025interleaved} with its Attention-driven Selection (ADS) mechanism and DyFo~\cite{li2025dyfo}, which utilizes Monte Carlo Tree Search (MCTS), have aimed to enhance the localization of relevant visual concepts by improving the discriminability of visual features and enabling dynamic focus adjustment. These efforts are complemented by approaches that explicitly model region-level visual relationships or incorporate structural constraints to promote more precise visual grounding. Despite these advances, such techniques have been primarily applied to enhance reasoning or generation capabilities, with limited integration into retrieval-oriented embedding learning. Our work bridges this gap by incorporating fine-grained visual understanding into multimodal representation learning, explicitly addressing subject-level alignment and visual neglect in retrieval contexts.

\section{Method}
Figure~\ref{fig:method} presents the overall architecture of the proposed SSA-ME framework, which introduces two core components: the Saliency-Guided Attention Alignment (SGA) module and the Saliency-Driven Feature Regeneration (SDR) module. Section~\ref{sec:eol} details the foundational multimodal embedding pipeline, while Sections~\ref{sec:sga} and~\ref{sec:sdr} elaborate on the underlying principles and training strategies of SGA and SDR, respectively. Finally, Section~\ref{sec:infer} describes the complete inference procedure.

\subsection{Multimodal Feature Extraction}
\label{sec:eol}
The SSA-ME framework is built upon three core components: a vision encoder for extracting hierarchical visual features, a vision projector that maps visual representations into the language space, and a language model for semantic understanding. Following the Explicit One-token Limitation (EOL) principle established in prior work~\cite{jiang2023scaling,jiang2024e5,liu2025lamra}, we employ distinct prompt templates for different input modalities. For visual inputs, we use the template \texttt{<image> <Instruction>}; for textual inputs, \texttt{<text> <Instruction>}. Multimodal inputs with interleaved images and text are processed using the extended template \texttt{<image\textsubscript{1}><text\textsubscript{1}>...<image\textsubscript{i}><text\textsubscript{j}> <Instruction>}. The content of \texttt{<Instruction>} is task-specific.

The feature extraction process begins with the vision encoder processing raw pixels to produce spatial visual features, which are then projected into the language model's embedding space. These projected visual tokens are concatenated with processed text tokens and fed into the language model's transformer. The final multimodal embedding is taken as the penultimate hidden state before the last token:
\begin{equation}
\mathcal{F}_{emb} = \phi(\mathcal{X}_{\text{visual}}, \mathcal{X}_{\text{text}}),
\end{equation}
where $\phi(\cdot)$ denotes the complete feature extraction process.
This unified extraction pipeline enables consistent representation learning across different input modalities, providing a solid foundation for subsequent saliency-aware processing.

\subsection{Saliency-Guided Attention Alignment}
\label{sec:sga}
To address the semantic alignment deviation problem identified in existing methods, we introduce a saliency-guided attention alignment mechanism that directs the model's focus toward semantically relevant visual regions. To accurately extract salient subjects, we design a prompt-based method using the query and its corresponding positive sample. As shown in Fig.~\ref{fig:method}, extracted subjects must be visually recognizable in the query image and semantically relevant to the positive sample, ensuring accurate visual priors during training.

Given a query pair (text $T_q$, image $I_q$) and a positive pair (text $T_t$, image $I_t$), we employ a LMM with a tailored instruction prompt $\mathcal{P}$ to identify salient regions:
\begin{equation}
\mathcal{S} = \{s_i\}_{i=1}^n = \text{LMM}(\mathcal{P}, T_q, I_q, T_t, I_t),
\end{equation}
where $\mathcal{S}$ represents the set of $n$ extracted salient subjects, and $\mathcal{P}$ is the specialized prompt template designed to leverage the LMM's perceptual capabilities.

For each extracted subject $s_i$, we employ the Segment Anything Model (SAM)~\cite{ravi2024sam} as a visual expert to generate a binary segmentation mask:
\begin{equation}
\mathcal{M}_i = \mathcal{V}(I_q, s_i), \quad \mathcal{M}_i \in \{0,1\}^{H \times W}.
\end{equation}
Here, $\mathcal{V}(\cdot,\cdot)$ denotes the SAM segmentation function, $H$ and $W$ represent the spatial dimensions of the input image, and $\mathcal{M}_i$ is a binary mask where values of 1 indicate pixels belonging to the salient subject.

To create biologically plausible attention distributions, we apply Gaussian smoothing~\cite{bergstrom2023gaussian} to convert binary masks into continuous saliency masks:
\begin{equation}
\mathcal{A}_i = \mathcal{M}_i \ast G_\sigma, \quad G_\sigma(x,y) = \frac{1}{2\pi\sigma^2}\exp\left(-\frac{x^2+y^2}{2\sigma^2}\right)
\end{equation}
where $\ast$ denotes the 2D convolution operation, $G_\sigma$ is the isotropic Gaussian kernel, and $\sigma$ controls the smoothing intensity. Larger $\sigma$ values produce broader attention distributions, while smaller values maintain sharper boundaries.

We ensure localization quality through a confidence validation mechanism. For each subject $s_i$, we compute:
\begin{equation}
\alpha_i = \kappa(\mathcal{E}_t(s_i), \mathcal{E}_v(I_i^s))
\end{equation}
where $I_i^s = \mathcal{M}_i \odot I_q$ is the subject region obtained via element-wise multiplication, $\mathcal{E}_t$ and $\mathcal{E}_v$ are CLIP's text and vision encoders, and $\kappa(\cdot,\cdot)$ computes cosine similarity. Subjects with $\alpha_i \geq \delta$ are retained, and the final saliency mask $\mathcal{A}$ is obtained by merging high-confidence maps via element-wise maximization:
\begin{equation}
\label{eq:clip}
\mathcal{A} = \max_{i: \, \alpha_i \geq \delta} \mathcal{A}_i.
\end{equation}

We introduce a saliency-guided loss $\mathcal{L}_{\text{SG}}$ to align the model's attention distributions $\{Q^{(l)}\}_{l=1}^{L}$ with the saliency mask $\mathcal{A}$:
\begin{equation}
\begin{split}
\mathcal{L}_{\text{SG}} &= \frac{1}{L}\sum_{l=1}^{L} \Bigl[ \beta \cdot D_{\text{KL}}\left(Q^{(l)} \parallel \mathcal{A}\right) \\
&\quad + (1-\beta) \cdot D_{\text{KL}}\left(\mathcal{A} \parallel Q^{(l)}\right) \Bigr]
\end{split}
\end{equation}
where $L$ is the total number of transformer layers, $Q^{(l)}$ represents the attention distribution at layer $l$, $D_{\text{KL}}$ denotes the Kullback-Leibler divergence measuring distribution discrepancy, and $\beta=0.5$ balances the bidirectional alignment.

\subsection{Saliency-Driven Feature Regeneration}
\label{sec:sdr}

To mitigate the visual modality neglect problem and enhance visual information utilization, we propose a feature regeneration mechanism that reinforces the model's reliance on salient visual content. This component addresses the tendency of multimodal models to underutilize visual information by dynamically emphasizing visually important regions.

The regeneration process begins with the attention patterns from the final transformer layer. The saliency map $Q \in \mathbb{R}^{H \times W}$ is flattened into a vector representation:
\begin{equation}
\hat{Q} = \text{Flatten}(Q) \in \mathbb{R}^{1 \times N}
\end{equation}
where $N = H \times W$ represents the total number of visual tokens, and $\text{Flatten}(\cdot)$ denotes the reshaping operation that preserves the sequential ordering of spatial locations.
The regenerated visual feature is computed through an attention-weighted aggregation:
\begin{equation}
\hat{\mathcal{F}_v} = \text{SoftMax}(\hat{Q}/\tau) \cdot \mathcal{F}_v
\end{equation}
where $\mathcal{F}_v \in \mathbb{R}^{N \times D}$ contains the visual features from the final layer, $D$ is the feature dimensionality, $\tau$ is a temperature parameter controlling the softmax concentration, and $\text{SoftMax}(\hat{Q}/\tau)$ produces a probability distribution over visual tokens. The resulting $\hat{\mathcal{F}_v} \in \mathbb{R}^{1 \times D}$ represents the saliency-weighted visual summary.

The final multimodal representation $\mathcal{F}_{\text{emb}} \in \mathbb{R}^{1 \times D}$ is formed by fusing the last token's feature $\mathcal{F}_{\text{last}}$ with the regenerated visual feature $\hat{\mathcal{F}_v}$. This fusion can be formulated as:
\begin{equation}
\mathcal{F}_{\text{emb}} = \mathcal{F}_{\text{last}} \oplus \hat{\mathcal{F}_v},
\end{equation}
where $\oplus$ denotes a feature fusion operator, such as concatenation followed by a linear projection or simple addition. In this paper, we use simple addition for fusion.

The training objective combines contrastive learning with saliency guidance:
\begin{equation}
    \mathcal{L}_{\text{total}} = \mathcal{L}_{Con} + \alpha \cdot \mathcal{L}_{\text{SG}}
    \label{eq:allloss}
\end{equation}
The contrastive loss $\mathcal{L}_{Con}$ follows the InfoNCE formulation~\cite{liu2025lamra}:
\begin{equation}
    \mathcal{L}_{Con} = -\frac{1}{B} \sum_{i=1}^{B} \log \frac{ \exp \left( \kappa ( \phi(q_i), \phi(c_i) ) / \tau \right) }{ \sum_{j=1}^{B} \exp \left( \kappa ( \phi(q_i), \phi(c_j) ) / \tau \right) }
\end{equation}
where $B$ is the batch size, $q_i = (T_q, I_q)$ denotes the $i$-th query pair, $c_i = (T_t, I_t)$ represents the corresponding candidate pair, and $\kappa(\cdot,\cdot)$ is the cosine similarity function.
The hyperparameter $\alpha$ controls the relative influence of saliency guidance.

\subsection{Inference Pipeline}
\label{sec:infer}
 Given an input query $q$ (which can be textual, visual, or multimodal) and a candidate pool $\{c_i\}_{i=1}^N$, the system executes the following steps: (1) Saliency Map Prediction: SSA-ME localizes salient subjects in the image based on the contextual information of the query text, generating a saliency distribution map Q; (2) Visual Feature Regeneration: Core visual features $\hat{\mathcal{F}}_v$ are adaptively retrieved from the image according to the saliency map Q and fused with the last token features $\mathcal{F}_{last}$ to form the final multimodal embedding $\mathcal{F}_{emb}$; (3) Similarity Computation and Ranking: Cosine similarity is computed between the query embedding and each candidate embedding, with the top-ranked samples selected as the final retrieval results.

\section{Experiment}
\subsection{Datasets and Metrics}
During training, we adhere to the VLM2Vec~\cite{jiang2024vlm2vec} benchmark and utilize the MMEB~\cite{jiang2024vlm2vec} dataset, which comprises 20 diverse datasets spanning four core multimodal tasks: classification, visual question answering, multimodal retrieval, and visual grounding. This comprehensive training corpus contains approximately 662K image-text pairs, encompassing both unimodal and multimodal inputs, ensuring the model's adaptability to a wide spectrum of multimodal challenges.
For evaluation, we assess all models on both in-distribution (the 20 test sets from MMEB) and out-of-distribution (16 additional test sets) benchmarks to thoroughly examine their multimodal embedding capabilities across various retrieval tasks. Following the standard evaluation protocol, we report the Precision@1 metric, which quantifies the proportion of correct matches among the top-ranked candidates for each dataset.

\subsection{Implementation Details}
Our framework employs Qwen2.5-VL-7B as the subject noun extraction model in SGA, with distinct strategies for different input modalities: for multimodal retrieval tasks (e.g., $(q^i,q^t)\to q^i$), relevant subjects are extracted exclusively from the textual query $q^t$, while for unimodal image inputs we leverage Qwen2.5-VL's multimodal understanding capability to identify the most representative visual subjects followed by precise segmentation using SAM to construct attention maps—an approach inspired by human cognitive patterns where visual search naturally focuses on salient objects. To ensure extraction quality, we implement a CLIP-based filtering mechanism that discards subjects with similarity scores below 0.2 ($\delta=0.2$ in Equation~\eqref{eq:clip}). The SGA component is configured with a temperature $\tau$ scaling parameter of 0.01 and balancing coefficient $\alpha=10$ to optimize the attention distribution in the last layer of the model. All our experiments were conducted on eight NVIDIA H800 GPUs with a batch size of 1024 and a learning rate of 1e-4. For Qwen2VL-2B, we trained for 5000 steps with a LoRA rank of 16, which took approximately 30 hours. For Qwen2VL-7B, we trained for 2000 steps with a LoRA rank of 8, which took approximately 35 hours.

\subsection{Main Results}
\label{subsec:main_res}

\begin{table*}[ht]
\centering
\caption{Results on the MMEB benchmark. The scores are averaged per meta-task, with IND and OOD denoting in-distribution and out-of-distribution test splits, respectively. VLM2Vec (Qwen2VL-2B) and VLM2Vec (Qwen2VL-7B) are our baselines.}
\label{tab:main_exp}
\resizebox{\textwidth}{!}{
\begin{tabular}{lccccccccc}
\toprule
\multirow{2}{*}{\textbf{Model}} & \multicolumn{4}{c}{\textbf{Per Meta-Task Score}} & & \multicolumn{3}{c}{\textbf{Average Score}} \\ 
\cmidrule(lr){2-5} \cmidrule(lr){7-9}
                       & Classification & VQA  & Retrieval & Grounding & & IND & OOD & Overall \\ \midrule
\# of datasets $\rightarrow$ & 10 & 10 & 12 & 4 & & 20 & 16 & 36 \\ \midrule

\multicolumn{9}{c}{\emph{Zero-shot on MMEB}} \\ \midrule
CLIP~\citep{radford2021learning}   & 42.8 & 9.1 &  53.0 &  51.8 &   &  37.1  &  38.7 &  37.8 \\
BLIP2~\citep{li2023blip}  & 27.0  &  4.2 & 33.9  & 47.0 &  &  25.3 &  25.1 & 25.2 \\
SigLIP~\citep{zhai2023sigmoid}  & 40.3  &  8.4 & 31.6  & 59.5 &  &  32.3 &  38.0 & 34.8 \\
OpenCLIP~\citep{cherti2023reproducible}  & 47.8  &  10.9 & 52.3  & 53.3 &  &  39.3 &  40.2 & 39.7 \\
UniIR (BLIP\_FF)~\citep{wei2023uniir} &  42.1 &	 15.0  &	60.1 & 	62.2  &	 & 44.7	&  40.4 & 	42.8 \\
UniIR (CLIP\_SF)~\citep{wei2023uniir} &  {44.3} & {16.2} & {61.8} & {65.3} & & {47.1}  & {41.7} &  {44.7}   \\
E5-V~\citep{jiang2024e5}  &  21.8 & 4.9 &  11.5 & 19.0 & & 14.9  &  11.5 & 13.3 \\
Magiclens~\citep{zhang2024magiclens}  &  38.8 &  8.3  &  35.4 &  26.0 &  & 31.0 & 23.7  & 27.8  \\ \midrule

\multicolumn{9}{c}{\emph{LMM-based Models $\le 2B$ (Trained on MMEB)}} \\ \midrule
CLIP~\cite{jiang2024vlm2vec}        &  55.2 &	19.7 & 	53.2 & 	62.2 &  & 47.6 &	42.8 &	45.4 \\
OpenCLIP~\cite{jiang2024vlm2vec}    & {56.0}	 &  {21.9} &  	{55.4} &  	{64.1} &  &	{50.5} &  	{43.1} &  	{47.2} \\ 
VLM2Vec (Phi-3.5-V)~\cite{jiang2024vlm2vec}                  & 54.8 & {54.9} & 62.3 & 79.5 & & 66.5 & 52.0 & 60.1  \\
LLaVE (Aquila-VL-2B)~\cite{lan2025llave}                  & 62.1 &{\textbf{60.2}} &65.2 &{\textbf{84.9}} & &69.4 &59.8 &65.2  \\

UniME (Phi-3.5-V)~\cite{gu2025breaking}                  & 54.8 & {55.9} & 64.5 &\textbf{81.8} & & 68.2 &52.7 &64.2  \\
UNITE$_{instruct}$~\cite{kong2025modality} & 63.2 & 55.9 &65.4 &75.6 & &65.8 &60.1 &63.3 \\
\rowcolor{gray!10}
VLM2Vec (Qwen2VL-2B)~\cite{jiang2024vlm2vec} & {59.0} &	{49.4}	 & {65.4}	 & {73.4} & &{66.0} & {52.6} & {60.1} \\
\rowcolor{black!10}
Ours (Qwen2VL-2B) & {\textbf{65.7}} &	{58.2}	 & {\textbf{68.9}}	 &  {77.5} & & 	{\textbf{70.9}} &  	{\textbf{59.8}} &  	{\textbf{66.0}} \\
\rowcolor{yellow!10}
$\Delta$ - baseline (2B)    &  +6.7 &   +8.8  & +3.5     &   +4.1   & &   +4.9   &  +7.2   & +5.9  \\

\midrule
\multicolumn{9}{c}{\emph{LMM-based Models $\geq 7B$ (Trained on MMEB)}} \\ 
\midrule
E5-V~\citep{jiang2024e5}  &  39.7 &10.8 & 39.4 & 60.2 & &34.2  &  33.4 & 33.9 \\
VLM2Vec (LLaVA-1.6-7B)~\cite{jiang2024vlm2vec} & {61.2} &	49.9	 & {67.4}	 &  {\textbf{86.1}} & & 	{67.5} &  	{57.1} &  	{62.9} \\
MMRet (LLaVA-Next-7B)~\cite{zhou2024megapairs} & {56.0} &59.4	 &69.9	 &83.6 & &68.0 &59.1 &64.1 \\
UniME (LLaVA-1.6-7B)~\cite{gu2025breaking}                  & 60.6 &52.9 &67.9 &85.1 & &68.4 &57.9 &66.6  \\

\rowcolor{gray!10}
VLM2Vec (Qwen2VL-7B)~\cite{jiang2024vlm2vec} & {62.6} &{57.8}	 & {69.9}	 & {81.7} & &{72.2} &{57.8} &{65.8} \\
\rowcolor{black!10}
Ours (Qwen2VL-7B) & {\textbf{67.4}} &	{\textbf{61.5}}	 & {\textbf{72.4}}	 &  {81.3} & & 	{\textbf{73.5}} &  	{\textbf{63.3}} &  	{\textbf{69.0}} \\
\rowcolor{yellow!10}
$\Delta$ - baseline  (7B)      &  +4.8 &   +3.7  & +2.5     &-0.4   & &   +1.3    &  +5.5    & +3.2  \\
\bottomrule
\end{tabular}
}
\end{table*}

\textbf{Comparison with other models on the MMEB benchmark.}
As demonstrated in table~\ref{tab:main_exp}, our proposed SSA-ME framework achieves state-of-the-art performance on the comprehensive MMEB benchmark. The results reveal significant improvements across all evaluation metrics compared to existing methods.
Specifically, our SSA-ME (Qwen2VL-2B) variant establishes new benchmarks in all four meta-tasks: classification (65.7\%), VQA (58.2\%), retrieval (68.9\%), and grounding (77.5\%). Notably, it achieves substantial gains of +6.7\% and +8.8\% over the strong VLM2Vec (Qwen2VL-2B) baseline in classification and VQA tasks, respectively.
In the overall evaluation, our method obtains remarkable improvements of +5.9\% overall score compared to baseline. More importantly, the model shows superior generalization capabilities with +4.9\% and +7.2\% gains on in-distribution (IND) and out-of-distribution (OOD) datasets, respectively, indicating robust performance across diverse data distributions.
Furthermore, when scaling to the 7B parameter configuration, SSA-ME (Qwen2VL-7B) achieves 69.0\% overall score, surpassing both MMRet by +4.9\% and VLM2Vec (Qwen2VL-7B) by +3.2\%. This consistent superiority across different model sizes validates the scalability and effectiveness of SSA-ME.
\begin{figure}[t]
  \centering
   \includegraphics[width=0.98\linewidth]{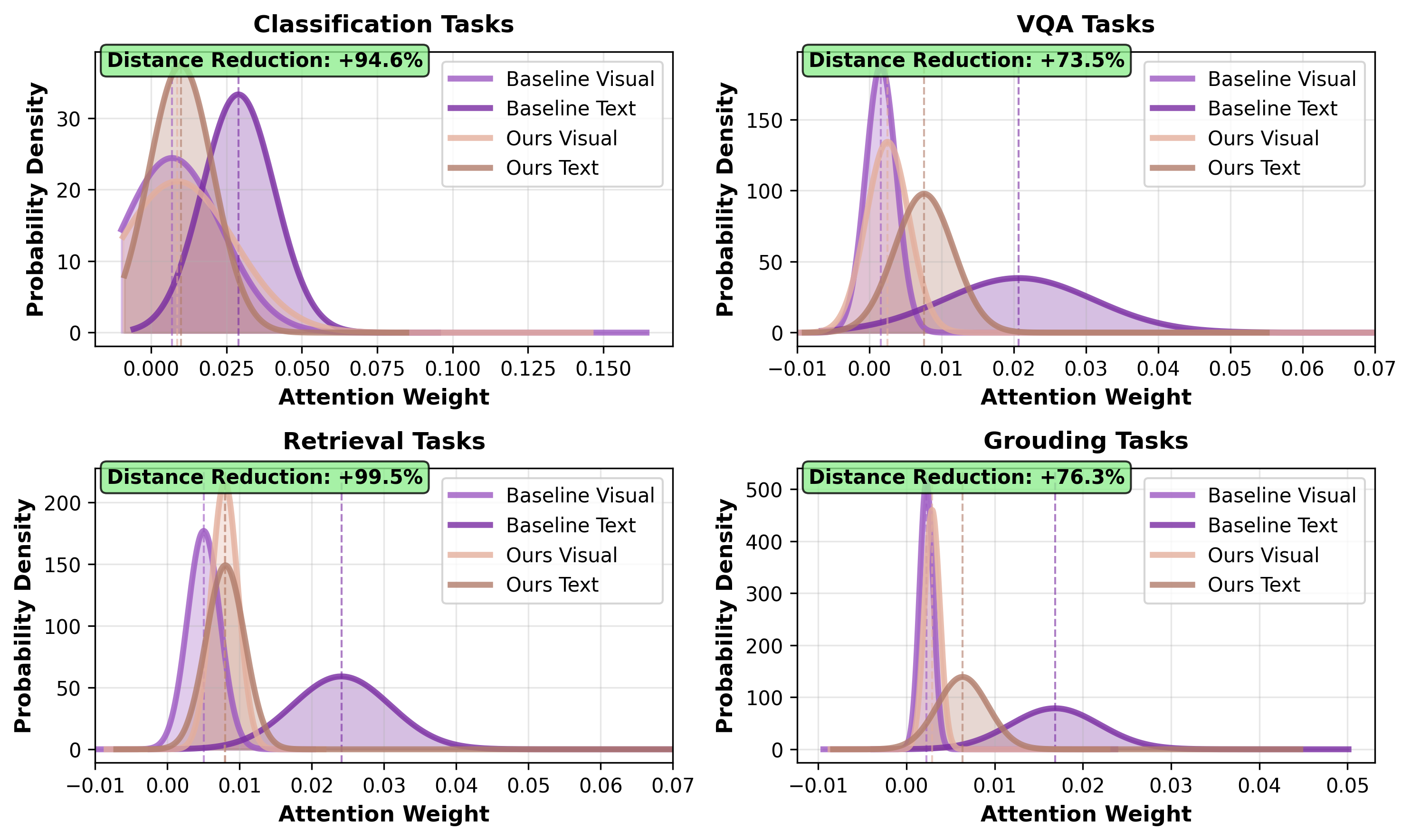}
   \caption{Qualitative comparison of attention weight distributions between visual and textual modalities for our method versus the VLM2Vec baseline.}
   \label{fig:dis}
\end{figure}
\\\textbf{Comparison of SSA-ME and baseline in visual-text modal attention distribution.}
As illustrated in Figure~\ref{fig:dis}, our method achieves remarkable balance between visual and textual attention weights across all four multimodal tasks. The probability density curves show that our approach significantly reduces the distribution distance between modalities compared to the baseline. 
The distance metric is calculated as:
\(D = \left| \mu_{\text{img}} - \mu_{\text{text}} \right|\), 
where $\mu_{\text{img}}$ and $\mu_{\text{text}}$ represent the mean attention weights of visual and textual modalities respectively. The percentage improvement is computed as:
\((D_{\text{baseline}} - D_{\text{ours}})/D_{\text{baseline}}.\)
We observe distance reductions of +94.6\% for classification, +73.5\% for VQA, +99.5\% for retrieval, and +76.3\% for grounding tasks. These results demonstrate that our saliency-driven framework effectively mitigates the visual modality neglect problem by bringing the attention distributions of both modalities closer to the ideal balanced state.

\subsection{Qualitative Results}
\begin{figure*}[ht]
  \centering
   \includegraphics[width=0.98\linewidth]{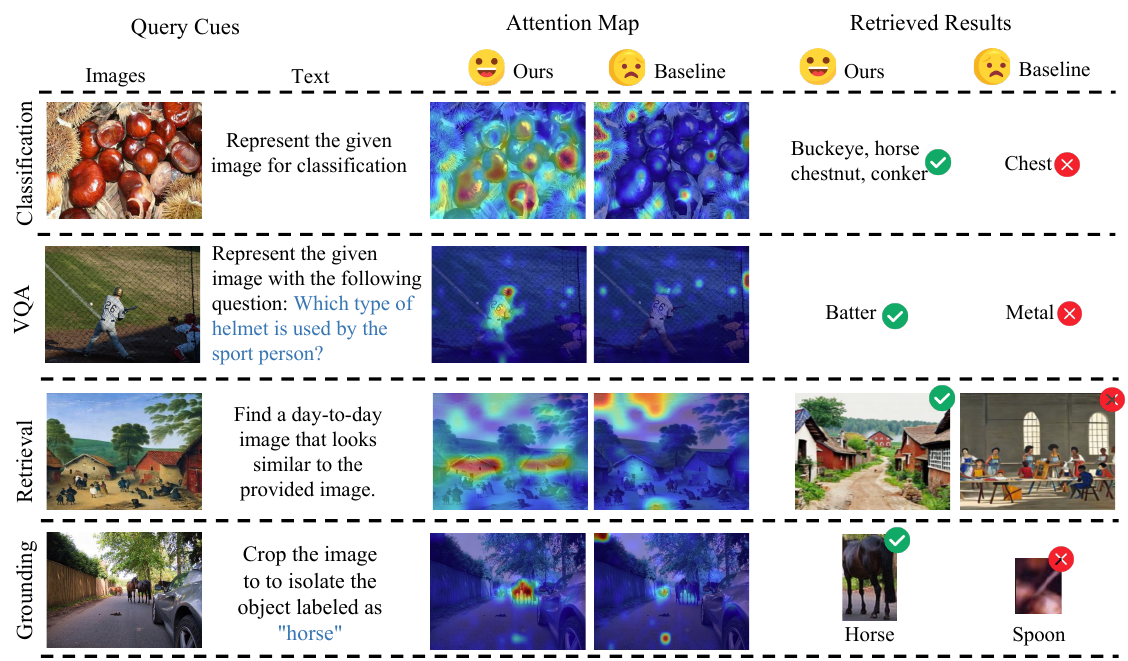}
   \caption{Demonstration of Selected Multimodal Retrieval Examples. The heatmaps visualize the attention distribution generated by the model when constructing embeddings for the input queries. As shown, SSA-ME exhibits significantly more precise localization of the target subjects compared to the baseline, leading to more accurate retrieval results.}
   \label{fig:false}
\end{figure*}
Figure~\ref{fig:false} presents a comprehensive qualitative analysis comparing our proposed SSA-ME framework with the baseline method across four multimodal retrieval tasks. 
In the \textbf{Classification} task, the query image containing chestnuts demonstrates distinct attention patterns between models. The baseline exhibits diffuse attention distribution across the entire image, while the proposed method shows more concentrated focus on the chestnut regions, leading to more precise feature localization.
For the \textbf{VQA} task (identifying helmet type), the attention maps reveal significant differences in spatial focus. Our method specifically concentrates on the baseball player's helmet area, whereas the baseline model exhibits noticeable attention shift towards image margins, resulting in incorrect classification.
In the \textbf{Retrieval} task, our approach accurately captures the architectural style of the house and land color characteristics, while the baseline model fails to focus on key elements, leading to erroneous retrieval results.
The \textbf{Grounding} task further demonstrates our method's capability to localize all horses in the image while identifying detailed equine features, achieving correct target retrieval. In contrast, the baseline only focuses on distant horses and misallocates attention to the upper-left corner and road surfaces, resulting in insufficient feature capture and incorrect target image retrieval.
Across all tasks, our saliency-driven attention mechanism consistently produces more focused and semantically relevant heatmaps, directly translating to improved retrieval performance. These qualitative results substantiate the quantitative improvements reported in Section~\ref{subsec:main_res}, confirming that SSA-ME effectively addresses semantic alignment deviation and visual modality neglect.

\subsection{Ablation Studies}
\textbf{Ablation study on different components.} Table~\ref{tab:ablation} presents a comprehensive ablation study evaluating the individual contributions of different components in our proposed SSA-ME framework on the MMEB benchmark. The ``Base'' configuration represents the VLM2Vec (Qwen2VL-2B) model trained solely with the standard InfoNCE loss.
The integration of the SGA component demonstrates substantial performance improvements, boosting the metrics to 70.7\% (IND), 58.4\% (OOD), and 65.3\% (Overall), corresponding to gains of +4.7\%, +5.8\%, and +5.2\% respectively. 
Notably, the SDR module alone also shows considerable improvements, achieving 70.4\% (IND), 57.9\% (OOD), and 64.9\% (Overall), with particularly strong OOD performance gains of +5.3\% over the base configuration. 
The complete SSA-ME framework, incorporating both SGA and SDR components, achieves the optimal performance of 70.9\% (IND), 59.8\% (OOD), and 66.0\% (Overall). The synergistic combination yields additional improvements of +1.4\% in OOD performance compared to the SGA-only configuration, demonstrating enhanced generalization capability. The superior OOD performance (59.8\% vs. 52.6\% in base) underscores the framework's robustness in handling distribution shifts.
\begin{table}[t]
\centering
\caption{Ablation study on different components.}
\resizebox{0.49\textwidth}{!}{%
\begin{tabular}{ccc|cccc|cc|c}
    \toprule
    {Base} & {SGA} & {SDR} & {CLS} & {VQA} & {RET} & {GD} & {IND} & {OOD}  &{Overall} \\
    \midrule
    $\checkmark$ & & &59.0 &49.4 &65.4 &73.4 & 66.0 & 52.6 & 60.1 \\
    $\checkmark$ & $\checkmark$ & &64.9 &57.3 &68.5 &76.7 & 70.7 & 58.4 & 65.3 \\
    $\checkmark$ & & $\checkmark$ &64.5 &57.5 &68.4 &73.4 &70.4 & 57.9 & 64.9 \\
    $\checkmark$ & $\checkmark$ & $\checkmark$ &\textbf{65.7} &\textbf{58.2} &\textbf{68.9} &\textbf{77.5} & \textbf{70.9} & \textbf{59.8} & \textbf{66.0} \\
    \bottomrule	
\end{tabular}%
}
\label{tab:ablation}
\end{table}
\begin{table}[t]
\centering
\caption{Impact of attention alignment layer position.}
\resizebox{0.45\textwidth}{!}{%
\begin{tabular}{c|cccc|cc|c}
    \toprule
    {Exp} & {CLS} & {VQA} & {RET} & {GD} & {IND} & {OOD}  &{Overall} \\
    \midrule
    Early &63.9 &57.1 &68.0 &74.9 &69.1  &58.9  &64.6  \\
    Middle &65.0 &57.5 &68.9 &73.8 & 70.3 &58.9 & 65.2 \\
    Late &\textbf{65.7} &\textbf{58.2} &68.9 &\textbf{77.5} & \textbf{70.9} & \textbf{59.8} & \textbf{66.0} \\
    All &64.3 &57.2 &\textbf{69.0} &74.2 &69.4 &59.5  &65.0  \\
    \bottomrule	
\end{tabular}%
}
\label{tab:ablation_attnpos}
\end{table}
\\\textbf{The influence of the position of the attention alignment layer.}
Table~\ref{tab:ablation_attnpos} presents the impact of applying our SGA mechanism at different layers of the model on retrieval performance. Here, ``Early'' denotes alignment applied to the first layer, ``Middle'' to a central layer, ``Late'' to the final layer, and ``All'' to all layers simultaneously. The results indicate that applying SGA to the final layer (``Late'') yields the best overall performance (66.0). In contrast, applying SGA to earlier or middle layers leads to noticeable performance degradation, with the ``Early'' configuration performing the worst (64.6 overall). Interestingly, aligning all layers (``All'') does not provide additional benefits compared to the ``Late'' configuration, suggesting that focusing on high-level semantic representations is most effective for saliency-guided alignment.
\begin{figure}[t]
  \centering
   \includegraphics[width=0.98\linewidth]{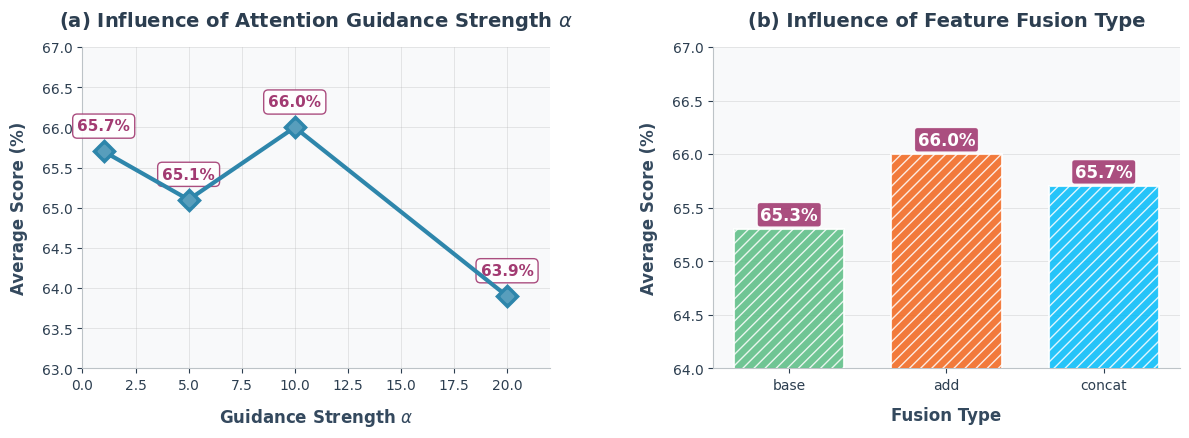}
   \caption{The impact of the guidance strength $\alpha$ and the feature fusion strategy on retrieval performance.}
   \label{fig:ablation}
\end{figure}
\\\textbf{Ablation study on model settings.}
Figure~\ref{fig:ablation} presents an ablation study analyzing the impact of the guidance strength $\alpha$ in $\mathcal{L}_{SG}$ (Eq.~\eqref{eq:allloss}) and the feature fusion strategy on retrieval performance. As shown in Fig.~\ref{fig:ablation} (a), performance peaks at $\alpha=10$ with an average score of 66.0\%, and degrades with stronger or weaker guidance (65.7\% at $\alpha=1$, 65.1\% at $\alpha=5$, and 63.9\% at $\alpha=20$), indicating the need for balanced saliency supervision. Fig.~\ref{fig:ablation} (b) compares fusion strategies: using only last token features (``base'') yields 65.3\%, while addition and concatenation achieve 66.0\% and 65.7\%, respectively, demonstrating that regenerated features significantly improve performance and addition provides the most effective integration. 
\begin{table}[t]
\centering
\caption{The impact of semantic relevance filtering.}
\resizebox{0.45\textwidth}{!}{%
\begin{tabular}{c|cccc|cc|c}
    \toprule
    {Exp} & {CLS} & {VQA} & {RET} & {GD} & {IND} & {OOD}  &{Overall} \\
    \midrule
    without filtering &64.0 & 57.6 & 68.9 & 74.9 & 70.5 & 58.2 & 65.0 \\
    with filtering &\textbf{65.7} &\textbf{58.2} &\textbf{68.9} &\textbf{77.5} & \textbf{70.9} & \textbf{59.8} & \textbf{66.0} \\
    \bottomrule	
\end{tabular}%
}
\label{tab:ablation-filter}
\end{table}
\\\textbf{The impact of semantic relevance filtering.} Table~\ref{tab:ablation-filter} evaluates the impact of semantic relevance filtering on the salient subject set $\mathcal{S}$ in SSA-ME. The results demonstrate that filtering yields consistent performance gains across most tasks. Notably, grounding accuracy improves from 74.9\% to 77.5\% and out-of-distribution generalization increases from 58.2\% to 59.8\%, while retrieval performance remains stable at 68.9\%. The overall accuracy rises from 65.0\% to 66.0\%, confirming that semantic filtering effectively eliminates irrelevant subjects and enhances the precision of saliency guidance. 
These improvements highlight the robustness of our filtering strategy in boosting multimodal retrieval through more reliable subject selection.

\section{Conclusion}
In this paper, we present a pioneering subject-level analysis that reveals two critical limitations in LMM-based embedding models trained with standard InfoNCE loss: semantic alignment deviation and visual modality neglect. To tackle these issues from a subject-centric perspective—a direction previously unexplored in multimodal retrieval—we introduce a simple yet effective Salient Subject-Aware Multimodal Embedding (SSA-ME) framework. SSA-ME incorporates two novel components: SGA and SDR, which work synergistically to enhance cross-modal semantic consistency and visual feature utilization by explicitly focusing on salient subject regions. Extensive experiments on the MMEB benchmark demonstrate that our subject-level approach significantly improves the model’s ability to capture salient visual-textual correspondences and achieve balanced multimodal representations. In the future, we will extend our approach to video-text retrieval and other complex multimodal tasks as we pursue more universal paradigms for embedding learning.

{
    \small
    \bibliographystyle{ieeenat_fullname}
    \bibliography{main}
}

\setcounter{section}{0}
\setcounter{figure}{0}
\setcounter{table}{0}
\renewcommand{\thefigure}{A\arabic{figure}}
\renewcommand{\thetable}{A\arabic{table}}
\clearpage
\setcounter{page}{1}
\maketitlesupplementary
\begin{abstract}
Section~\ref{sec:prompt} details the prompt engineering strategy for salient subject extraction and presents representative examples of the extracted subjects alongside their corresponding segmentation masks. Section~\ref{sec:stat_on_ss} provides a statistical analysis of the spatial distribution patterns of salient subjects within images, accompanied by word cloud visualizations that characterize the most frequent subject nouns across the four multimodal task categories. Section~\ref{sec:performance} presents a comprehensive performance comparison between our proposed method and existing approaches, reporting detailed evaluation metrics across all 36 datasets of the MMEB benchmark. Section~\ref{sec:number_vt} presents an ablation study on the number of visual tokens regenerated in the Saliency-Driven Feature Regeneration (SDR) module.
\end{abstract}

\section{Prompting for Salient Subjects}
\label{sec:prompt}
\begin{figure}[t]
  \centering
   \includegraphics[width=0.98\linewidth]{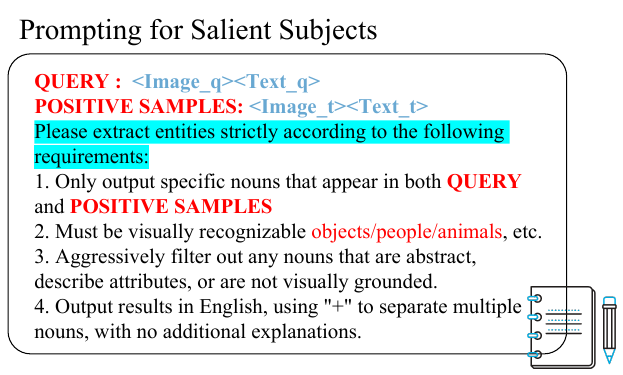}
   \caption{Prompt engineering pipeline for salient subject extraction.}
   \label{fig:prompt_pipeline}
\end{figure}

Figure~\ref{fig:prompt_pipeline} illustrates the prompt engineering pipeline for salient subject extraction in our SSA-ME framework. The process begins by constructing a structured input containing the query pair $(T_q, I_q)$ and positive sample pairs $(T_t, I_t)$, formatted as a unified sequence following the template: \texttt{QUERY:<Image\_q><Text\_q>; POSITIVE SAMPLES: <Image\_t><Text\_t>...}. This input is processed by a Large Multimodal Model (LMM) using a carefully designed instructional prompt $\mathcal{P}$ that specifies three key constraints: (1) only nouns appearing in both the query and positive samples are retained, (2) all entities must be visually recognizable, and (3) abstract or non-visually-grounded concepts are aggressively filtered out.

The LMM generates an initial set of salient subjects $\mathcal{S} = \{s_i\}_{i=1}^n$ satisfying the cross-sample co-occurrence and visual-grounding requirements. Each subject $s_i$ is then validated using a visual expert model to ensure semantic consistency between textual descriptions and corresponding image regions. The final output is a refined set of visually-grounded salient subjects that serve as the foundation for subsequent saliency-guided attention alignment and feature regeneration modules. This prompt-based extraction mechanism enables precise identification of semantically relevant visual concepts while effectively suppressing noise from irrelevant or abstract entities.

\begin{figure*}[t]
  \centering
   \includegraphics[width=0.98\linewidth]{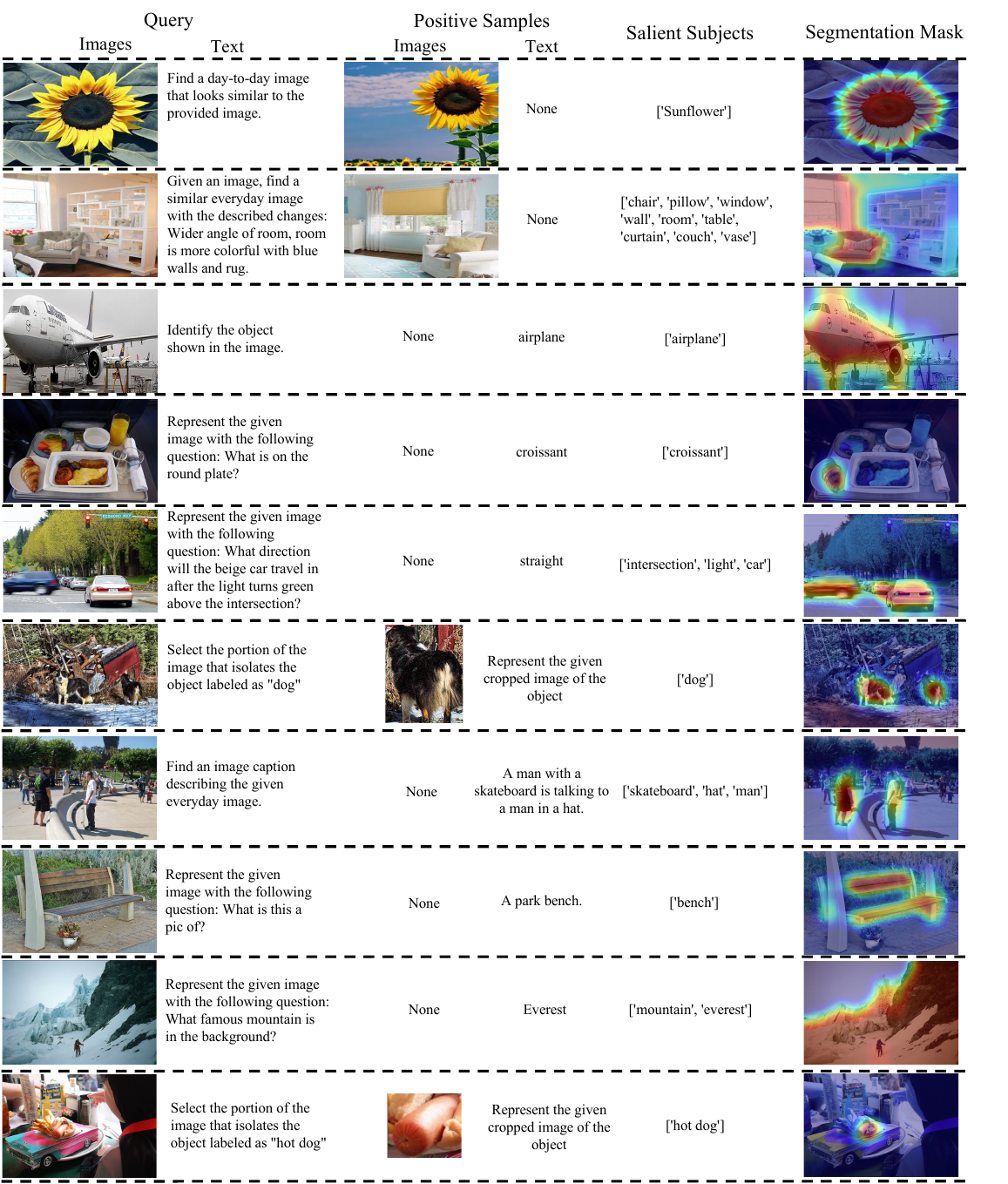}
   \caption{Examples of salient subject extraction results alongside their corresponding segmentation masks.}
   \label{fig:showmask}
\end{figure*}

Figure~\ref{fig:showmask} presents representative examples of salient subject extraction results alongside their corresponding segmentation masks. The visualization clearly demonstrates the effectiveness of our carefully designed extraction pipeline in accurately localizing salient regions within query images by leveraging semantic guidance from positive samples.
As shown in the figure~\ref{fig:showmask}, our method successfully identifies semantically meaningful subjects across diverse query types. For instance, when queried with \textit{``Find a day-to-day image that looks similar to the provided image"} alongside a sunflower image, our pipeline correctly identifies \textit{``Sunflower"} as the salient subject. Similarly, for complex queries involving directional reasoning about vehicles or scene understanding, the extraction maintains precise focus on relevant elements like \textit{``intersection"}, \textit{``light"}, and \textit{``car"}.
This subject-level grounding enables the model to filter out irrelevant visual noise and concentrate on task-critical information, ultimately leading to embeddings with substantially improved signal-to-noise ratio. The consistent performance across various query types—from object identification to relational reasoning—validates the robustness and generalizability of our salient subject extraction framework.

\section{Statistics on Salient Subjects}
\label{sec:stat_on_ss}

\begin{figure}[t]
  \centering
   \includegraphics[width=0.98\linewidth]{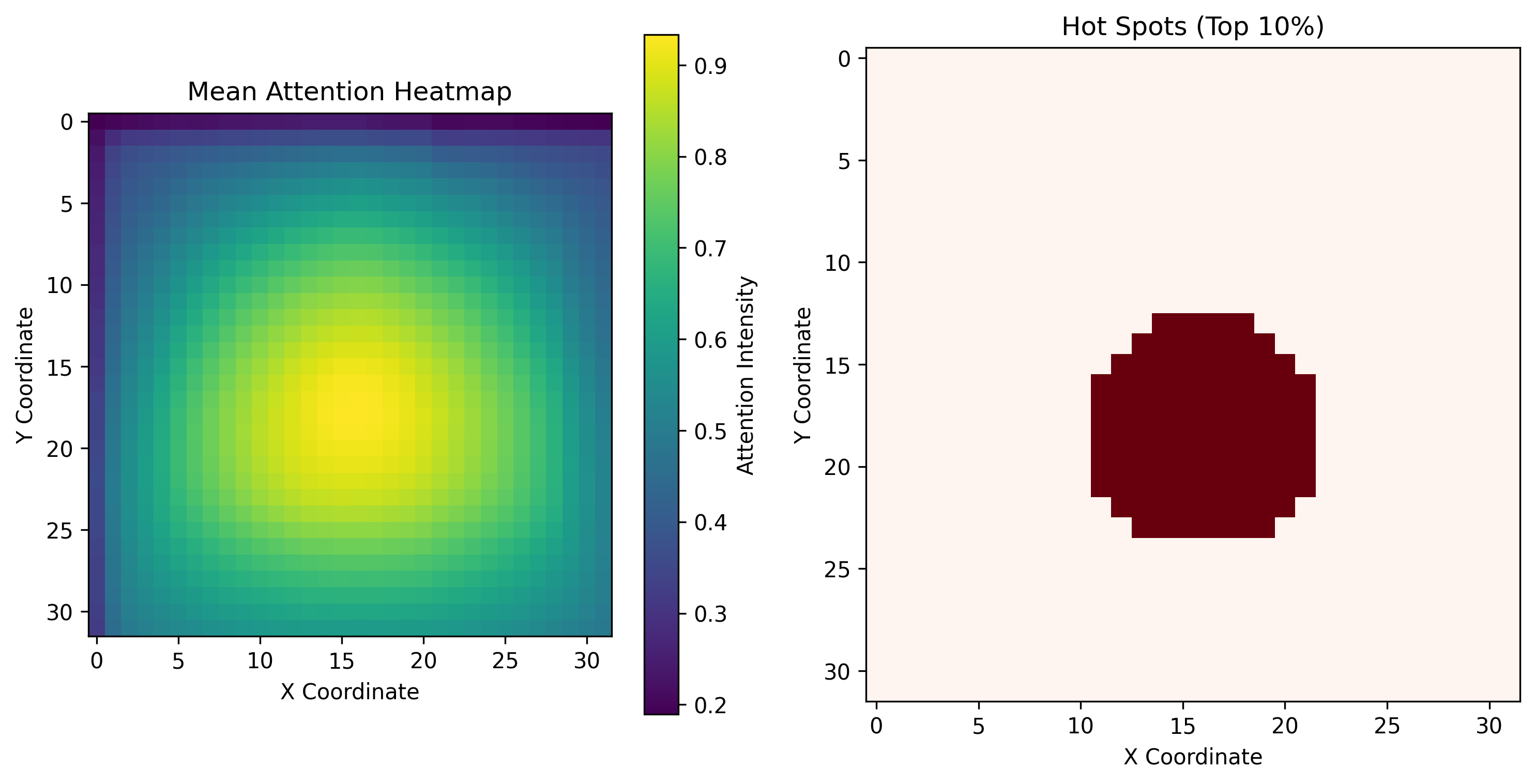}
   \caption{Distribution of Salient Subjects. The figure presents a quantitative analysis of spatial attention distribution for extracted salient subjects, combining a mean attention heatmap (left) and top 10\% hot spots (right). Both visualizations reveal a strong central concentration pattern, indicating that the majority of salient subjects are predominantly located in the central image regions.}
   \label{fig:attnmap}
\end{figure}

The spatial analysis of salient subject distribution reveals a significant central concentration bias in attention patterns. As shown in the mean attention heatmap, attention intensity peaks around the central coordinates with values reaching 0.9 (bright yellow), gradually decreasing toward the image boundaries to approximately 0.2 (dark purple). This radial gradient pattern is further confirmed by the hot spots analysis, where the top 10\% most attended regions form a compact cluster centered in the middle of the coordinate space.
The consistent central concentration across both visualizations suggests that our extraction methodology effectively identifies the most semantically relevant subjects, which tend to occupy prominent central positions in natural images. This distribution pattern aligns with typical photographic composition principles where primary subjects are often placed near the image center. 
This spatial characterization provides valuable insights for optimizing attention mechanisms in multimodal models, particularly for tasks requiring precise subject localization and background suppression.

\begin{figure}[ht]
  \centering
   \includegraphics[width=0.98\linewidth]{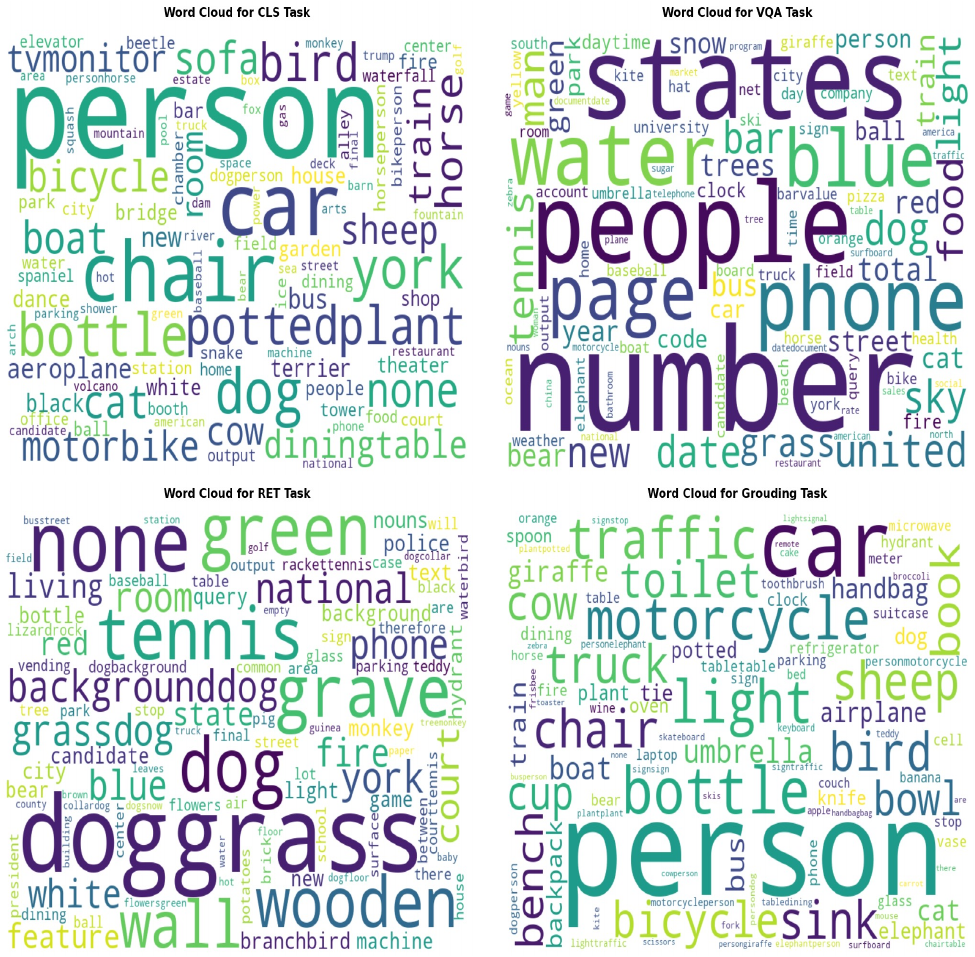}
   \caption{Word clouds of salient subjects extracted from four tasks.}
   \label{fig:wordcloud}
\end{figure}

The word clouds in Figure ~\ref{fig:wordcloud} provide a quantitative overview of the salient subjects identified by our extraction pipeline across four distinct multimodal tasks: Classification (CLS), Visual Question Answering (VQA), Retrieval (RET), and Visual Grounding. A clear divergence in the distribution of salient concepts across tasks can be observed, underscoring the task-dependent nature of visual semantics. For instance, the CLS task is dominated by concrete object categories such as \textit{``person''}, \textit{``car''}, and \textit{``bird''}, reflecting its focus on object recognition. In contrast, the VQA task features more abstract concepts like \textit{``states''} and \textit{``number,''} indicating a stronger emphasis on perceptual and reasoning-oriented information. This task-specific distribution validates that our method effectively captures semantically relevant subjects tailored to different multimodal objectives. The high-frequency keywords further demonstrate the capability of our approach to filter out irrelevant visual noise and concentrate on the most salient entities, which is fundamental for learning discriminative and high signal-to-noise ratio embeddings.

\section{Overall Performance Comparison}
\label{sec:performance}

Table~\ref{tab:main_exp_per_task} presents a comprehensive performance comparison across 36 diverse vision-language tasks, where {VLM2Vec} employs {LLaVA-1.6-7B} as its base model, while our proposed {SSA-ME} method is evaluated in two configurations: {SSA-ME (2B)} using {Qwen2VL-2B} and {SSA-ME (7B)} using {Qwen2VL-7B} as base models respectively. The results demonstrate that the superior and robust performance of our proposed SSA-ME framework. Notably, our method establishes new state-of-the-art results in the challenging Unified Multimodal Retrieval benchmark. The consistent superiority of SSA-ME across all task categories---Classification (10 tasks), VQA (10 tasks), Retrieval (12 tasks), and Visual Grounding (4 tasks)---validates that explicitly modeling salient subjects effectively addresses semantic alignment deviation and visual modality neglect, leading to more balanced and semantically consistent embeddings. The performance gap is particularly pronounced in complex tasks such as DocVQA, where SSA-ME (7B) achieves 92.2 compared to VLM2Vec's 52.0, demonstrating its enhanced capability for fine-grained visual-textual understanding.

\begin{table*}[t]
\centering
\caption{The detailed results of the baselines and our SSA-ME on MMEB, which includes 20 in-distribution datasets and 16 out-of-distribution datasets. The out-of-distribution datasets are highlighted with a yellow background in the table. 
}
\resizebox{0.98\textwidth}{!}{
\begin{tabular}{lcccccccccc}
\toprule
\rowcolor{gray!30}  & \textbf{CLIP} & \textbf{OpenCLIP} & \textbf{SigLIP} & \textbf{BLIP2} & \textbf{MagicLens} & \textbf{E5-V} & \textbf{UniIR} & \textbf{VLM2Vec} & \textbf{SSA-ME (2B)} & \textbf{SSA-ME (7B)}\\
\midrule
\rowcolor{orange!30} \textbf{Classification (10 tasks)} & & & & & & & & & &\\
ImageNet-1K          & 55.8 & 63.5 & 45.4 & 10.3 & 48.0 & 9.6 & 58.3 & 74.5 & 81.9 & 84 \\
N24News              & 34.7 & 38.6 & 13.9 & 36.0 & 33.7 & 23.4 & 42.5 & 80.3 & 79.1 & 80.2 \\
HatefulMemes         & 51.1 & 51.7 & 47.2 & 49.6 & 49.0 & 49.7 & 56.4 & 67.9 & 62.8 & 68.4 \\
VOC2007              & 50.7 & 52.4 & 64.3 & 52.1 & 51.6 & 49.9 & 66.2 & 91.5 & 86.4 & 84.5 \\
SUN397               & 43.4 & 68.8 & 39.6 & 34.5 & 57.0 & 33.1 & 63.2 & 75.8 & 76.5 & 75.1 \\
\rowcolor{yellow!15} Place365  & 28.5 & 37.8 & 20.0 & 21.5 & 31.5 & 8.6 & 36.5 & 44.0 & 40.9 & 43.6 \\
\rowcolor{yellow!15} ImageNet-A & 25.5 & 14.2 & 42.6 & 3.2  & 8.0  & 2.0 & 9.8 & 43.6 & 48.7 & 53.1 \\
\rowcolor{yellow!15} ImageNet-R & 75.6 & 83.0 & 75.0 & 39.7 & 70.9 & 30.8 & 66.2 & 79.8 & 88.7 & 88 \\
\rowcolor{yellow!15} ObjectNet  & 43.4 & 51.4 & 40.3 & 20.6 & 31.6 & 7.5 & 32.2& 39.6 & 65.5 & 66.2 \\
\rowcolor{yellow!15} Country-211 & 19.2 & 16.8 & 14.2 & 2.5  & 6.2  & 3.1 & 11.3 & 14.7 & 26.5 & 30.8 \\
\textit{All Classification} & 42.8 & 47.8 & 40.3 & 27.0 & 38.8 & 21.8 & 44.3 & 61.2 & 65.7 & 67.4 \\
\midrule

\rowcolor{blue!30} \textbf{VQA (10 tasks)} & & & & & & & & & & \\
OK-VQA               & 7.5  & 11.5 & 2.4  & 8.7  & 12.7 & 8.9 & 25.4 & 69.0 & 60.6 & 60.7 \\
A-OKVQA              & 3.8  & 3.3  & 1.5  & 3.2  & 2.9  & 5.9 & 8.8 & 54.4 & 51.4 & 52.1 \\
DocVQA               & 4.0  & 5.3  & 4.2  & 2.6  & 3.0  & 1.7 & 6.2 & 52.0 & 91.5 & 92.2 \\
InfographicsVQA      & 4.6  & 4.6  & 2.7  & 2.0  & 5.9  & 2.3 & 4.6 & 30.7& 59.7 & 70.3 \\
ChartQA              & 1.4  & 1.5  & 3.0  & 0.5  & 0.9  & 2.4 & 1.6 & 34.8 & 50.8 & 57.9 \\
Visual7W             & 4.0  & 2.6  & 1.2  & 1.3  & 2.5  & 5.8 & 14.5 & 49.8 & 54.1 & 54.6 \\
\rowcolor{yellow!15} ScienceQA  & 9.4  & 10.2 & 7.9  & 6.8  & 5.2  & 3.6 & 12.8 & 42.1 & 36.9 & 46.5 \\
\rowcolor{yellow!15} VizWiz    & 8.2  & 6.6  & 2.3  & 4.0  & 1.7  & 2.6  & 24.3 & 43.0 & 46.1 & 43.7 \\
\rowcolor{yellow!15} GQA        & 41.3 & 52.5 & 57.5 & 9.7  & 43.5 & 7.8 & 48.8 & 61.2 & 54.7 & 60.5 \\
\rowcolor{yellow!15} TextVQA    & 7.0  & 10.9 & 1.0  & 3.3  & 4.6  & 8.2 & 15.1 & 62.0 & 76 & 77 \\
\textit{All VQA}      & 9.1  & 10.9 & 8.4  & 4.2  & 8.3  & 4.9 & 16.2 & 49.9 & 58.2 &61.5 \\
\midrule

\rowcolor{green!30} \textbf{Retrieval (12 tasks)} & & & & & & & & & & \\
VisDial              & 30.7 & 25.4 & 21.5 & 18.0 & 24.8 & 9.2 & 42.2 & 80.9 & 81.7 & 85.4 \\
CIRR                 & 12.6 & 15.4 & 15.1 & 9.8  & 39.1 & 6.1 & 51.3 & 49.9 & 54.2 & 59.6 \\
VisualNews\_t2i      & 78.9 & 74.0 & 51.0 & 48.1 & 50.7 & 13.5 & 74.3 & 75.4 & 75.5 & 79.6 \\
VisualNews\_i2t      & 79.6 & 78.0 & 52.4 & 13.5 & 21.1 & 8.1 & 76.8 & 80.0 & 78.3 & 83.7 \\
MSCOCO\_t2i          & 59.5 & 63.6 & 58.3 & 53.7 & 54.1 & 20.7 & 68.5 & 75.7 & 76.9 & 78.4 \\
MSCOCO\_i2t          & 57.7 & 62.1 & 55.0 & 20.3 & 40.0 & 14.0 & 72.1 & 73.1 & 73.7 & 74.1 \\
NIGHTS               & 60.4 & 66.1 & 62.9 & 56.5 & 58.1  & 4.2 & 66.2 & 65.5 & 68.2 & 68.4 \\
WebQA                & 67.5 & 62.1 & 58.1 & 55.4 & 43.0 & 17.7 & 89.6 & 87.6 & 88.7 & 90.9 \\
\rowcolor{yellow!15} FashionIQ  & 11.4 & 13.8 & 20.1 & 9.3  & 11.2 & 2.8 & 40.2 & 16.2 & 18.5 & 21.1 \\
\rowcolor{yellow!15} Wiki-SS-NQ & 55.0 & 44.6 & 55.1 & 28.7 & 18.7 & 8.6 & 12.2 & 60.2 & 63.9 & 67.5 \\
\rowcolor{yellow!15} OVEN       & 41.1 & 45.0 & 56.0 & 39.5 & 1.6  & 5.9 & 69.4 & 56.5 & 64.5 & 70.6 \\
\rowcolor{yellow!15} EDIS       & 81.0 & 77.5 & 23.6 & 54.4 & 62.6 & 26.8 & 79.2 & 87.8 & 82.4 & 89.3 \\
\textit{All Retrieval} & 53.0 & 52.3 & 31.6 & 33.9 & 35.4 & 11.5 & 61.8 & 67.4 & 68.9 &72.4 \\
\midrule

\rowcolor{purple!30} \textbf{Visual Grounding (4 tasks)} & & & & & & & & & & \\
MSCOCO         & 33.8 & 34.5 & 46.4 & 28.9 & 22.1 & 10.8 & 46.6 & 80.6 & 66.6 & 70.1 \\
\rowcolor{yellow!15} RefCOCO  & 56.9 & 54.2 & 70.8 & 47.4 & 22.8 & 11.9 & 67.8 & 88.7 & 84.6 & 89.6 \\
\rowcolor{yellow!15} RefCOCO-matching  & 61.3 & 68.3 & 50.8 & 59.5 & 35.6 & 38.9 & 62.9 & 84.0 & 87 & 90.5 \\
\rowcolor{yellow!15} Visual7W-pointing & 55.1 & 56.3 & 70.1 & 52.0 & 23.4 & 14.3 & 71.3 & 90.9 & 71.9 & 75 \\
\textit{All Visual Grounding} & 51.8 & 53.3 & 59.5 & 47.0 & 26.0 & 19.0 & 65.3 & 86.1 &77.5 &81.3 \\
\midrule

\rowcolor{cyan!15} \textbf{Final Score (36 tasks)} & & & & &  & & & & & \\
All                  & 37.8 & 39.7 & 34.8 & 25.2  & 27.8 & 13.3 & 44.7 & 62.9 &66.0 & 69.0\\
All IND       & 37.1 & 39.3 & 32.3 & 25.3  & 31.0 & 14.9 & 47.1 & 67.5 &70.9 & 73.5 \\
All OOD       & 38.7 & 40.2 & 38.0 & 25.1  & 23.7 &  11.5 & 41.7 & 57.1 &59.8 &63.3 \\

\bottomrule
\end{tabular}
}
\label{tab:main_exp_per_task}
\end{table*}

\section{Impact of the Number of Regenerated Visual Tokens}
\label{sec:number_vt}
\begin{table}[t]
\centering
\caption{Impact of attention alignment layer position.}
\resizebox{0.45\textwidth}{!}{%
\begin{tabular}{c|cccc|cc|c}
    \toprule
    {Exp} & {CLS} & {VQA} & {RET} & {GD} & {IND} & {OOD}  &{Overall} \\
    \midrule
    Top-1 &64.0 &56.2 &54.3 &29.2 &60.9  &46.9  &54.7  \\
    Top-10 &65.6 &58.0 &68.1 &77.1 & 70.5 &59.4 & 65.6 \\
    Top-50 &65.9 &58.1 &68.9 &77.3 & 70.9 & 59.8 & 66.0 \\
    All &\textbf{65.7} &\textbf{58.2} &\textbf{68.9} &\textbf{77.5} & \textbf{70.9} & \textbf{59.8} & \textbf{66.0} \\
    \bottomrule	
\end{tabular}%
}
\label{tab:token_ablation}
\end{table}
Table~\ref{tab:token_ablation} presents an ablation study on the impact of the number of regenerated visual tokens, selected based on saliency scores, on multimodal performance. The results reveal a clear positive correlation: performance across all metrics (CLS, VQA, RET, GD, IND, OOD, and Overall) improves substantially as the number of regenerated tokens $k$ increases from 1 to 50. Notably, the most significant performance leap occurs when expanding from Top-1 to Top-10 tokens, with the Overall score surging from 54.7 to 65.6. This suggests that regenerating features for a minimal set of tokens (Top-1) is insufficient, and capturing a broader set of salient visual cues is critical. Performance saturates beyond Top-50, as indicated by the marginal gains when using all tokens (Overall: 66.0). This saturation effect demonstrates that our saliency-driven regeneration effectively prioritizes the most informative tokens, achieving near-optimal performance without the need to process the entire token set, thereby maintaining a favorable balance between model efficacy and computational efficiency.

\end{document}